\documentclass[letterpaper]{article} 
\usepackage{aaai2027}  
\usepackage[hyphens]{url}  
\usepackage{graphicx} 
\usepackage{natbib}  
\usepackage{caption} 
\usepackage{booktabs}
\usepackage{amsmath}
\usepackage{tikz}
\usetikzlibrary{shapes.geometric, arrows.meta, positioning}
\usepackage{multirow}

\nocopyright

\title{Multi-Label Topic Assignment via LLM Distillation: A Comparative Analysis of Generative vs. Discriminative Student Models}

\author{
    Sourabh Kasliwal\textsuperscript{\rm 1},
    Shubhranshu Singh\textsuperscript{\rm 2}
}
\affiliations{
    \textsuperscript{\rm 1}eBay Inc., Bellevue, WA\\
    \textsuperscript{\rm 2}eBay Inc., Portland, OR\\
    sokasliwal@ebay.com, shubhrasingh@ebay.com
}

\begin{document}
\maketitle
\begin{abstract}
Multi-label topic assignment for user-generated content (UGC)—including product reviews and buyer-seller conversations—poses unique scalability challenges in large-scale e-commerce due to informal language, extreme label sparsity, and rapidly evolving taxonomies. While utilizing Large Language Models (LLMs) as labeling oracles to distill ground-truth data has emerged as an industry standard to bypass prohibitive manual annotation costs, determining the optimal, low-latency architecture for the resulting student models remains an open challenge. To address this, we conduct a comprehensive evaluation across Small Language Model (SLM) parameter scales (1B, 4B, and 8B) and architectural paradigms (causal generative versus bidirectional discriminative). Comparing generative text-to-label classifiers against discriminative baselines (DeBERTa-V3 and ModernBERT), our analysis reveals a crucial data-dependent trade-off: while discriminative models outperform ultra-lightweight generative models on structured product reviews, even the smallest 1B generative model surpasses discriminative baselines on complex, multi-turn conversational data. Furthermore, generative models maintain robust performance under massive label-set expansion (up to 112 topics) and severe long-tail distributions, whereas discriminative baselines suffer a 35\% drop in Macro-F1 at scale. Finally, we detail the successful production deployment of these optimized models across both product review and conversational domains, demonstrating strict latency compliance and tangible business impact at a global marketplace scale.\end{abstract}

\section{Introduction}
\label{sec:introduction}

At a global e-commerce marketplace, the ability to categorize and derive insights from user-generated content (UGC) is essential for maintaining platform health and buyer trust. With millions of active listings across thousands of categories---from automotive parts to collectible electronics---topic assignment is a critical driver for numerous business functions. These range from identifying risk, fraudulent activity, or user intent in buyer-seller messaging to generating structured review summaries that help buyers navigate complex purchasing decisions.

However, textual user-generated content in a marketplace ecosystem is uniquely challenging. Unlike standardized product descriptions, buyer-seller conversations and product reviews are often characterized by extreme linguistic variety, platform-specific jargon, and informal syntax. Traditional multi-label classification methods often struggle with the noise and sparsity inherent in this data \citep{llm_multilabel2025}. Furthermore, the sheer scale of the dataset makes manual labeling for every new niche or category extremely inefficient and cost-prohibitive. Recently, leveraging high-capacity Large Language Models (LLMs) as labeling oracles to synthesize training data for smaller ``student'' models has become a standard industry practice to reduce these manual annotation costs \citep{generativeai_multilabel2024}\citep{llm_multilabel2025}.

While LLM powered distillation solves the data bottleneck, determining the optimal architecture and parameter scale for the resulting student models remains an open engineering challenge. Production systems must balance inference latency against classification fidelity. Furthermore, it is unclear whether traditional discriminative boundaries or modern autoregressive generative decoding are better suited for the varying linguistic complexities of marketplace text---such as the difference between single-author product reviews and multi-turn, multi-author dialogues.

To address this, we present a comprehensive comparative analysis of multi-label classification architectures within a production Teacher-Student distillation framework. This paper makes the following core contributions:

\begin{itemize}
    \item \textbf{Empirical Comparison of Architectures and Scales:} We conduct a rigorous evaluation comparing traditional discriminative classifiers DeBERTa-V3\citep{DeBERTaV3} and ModernBERT\citep{warner2024smarterbetterfasterlonger} against generative text-to-label Small Language Models (SLMs) across varying parameter scales (1B, 4B, and 8B), assessing their fidelity to distilled oracle labels.

    \item \textbf{Discovery of Data-Dependent Trade-offs \& Taxonomy Robustness:} We reveal a critical performance divergence based on data complexity: while discriminative models outperform ultra-lightweight generative models on structured product reviews, even the smallest 1B generative model surpasses discriminative baselines on multi-turn conversational data. Additionally, we demonstrate that generative models maintain performance under massive label-set expansion (up to 112 topics), whereas discriminative baselines suffer a severe 35\% performance collapse.

    \item \textbf{Production Deployment and Business Impact:} Grounded in our comparative findings, we detail the production deployment of optimized generative multi-label models across multiple domains, demonstrating latency compliance and tangible business impact at scale.
\end{itemize}

The remainder of this paper is organized as follows: Section \ref{sec:methodology} discusses our methodology including the Teacher-Student framework, discriminative vs. generative architectures, and the prompt engineering template for LLM-based annotation. Section \ref{sec:experimental-setup} details dataset creation and topic classification challenges across multiple types of datasets(product reviews and buyer-seller conversations). Section \ref{sec:results} presents the experimental results and comparative analysis. Section \ref{sec:impact} details the production integration and its impact on review quality, followed by Section \ref{sec:conclusion}, which outlines future directions.

\section{Methodology: Scalable Multi-Label Classification}
\label{sec:methodology}

\begin{figure}[h]
\centering
\includegraphics[width=\columnwidth]{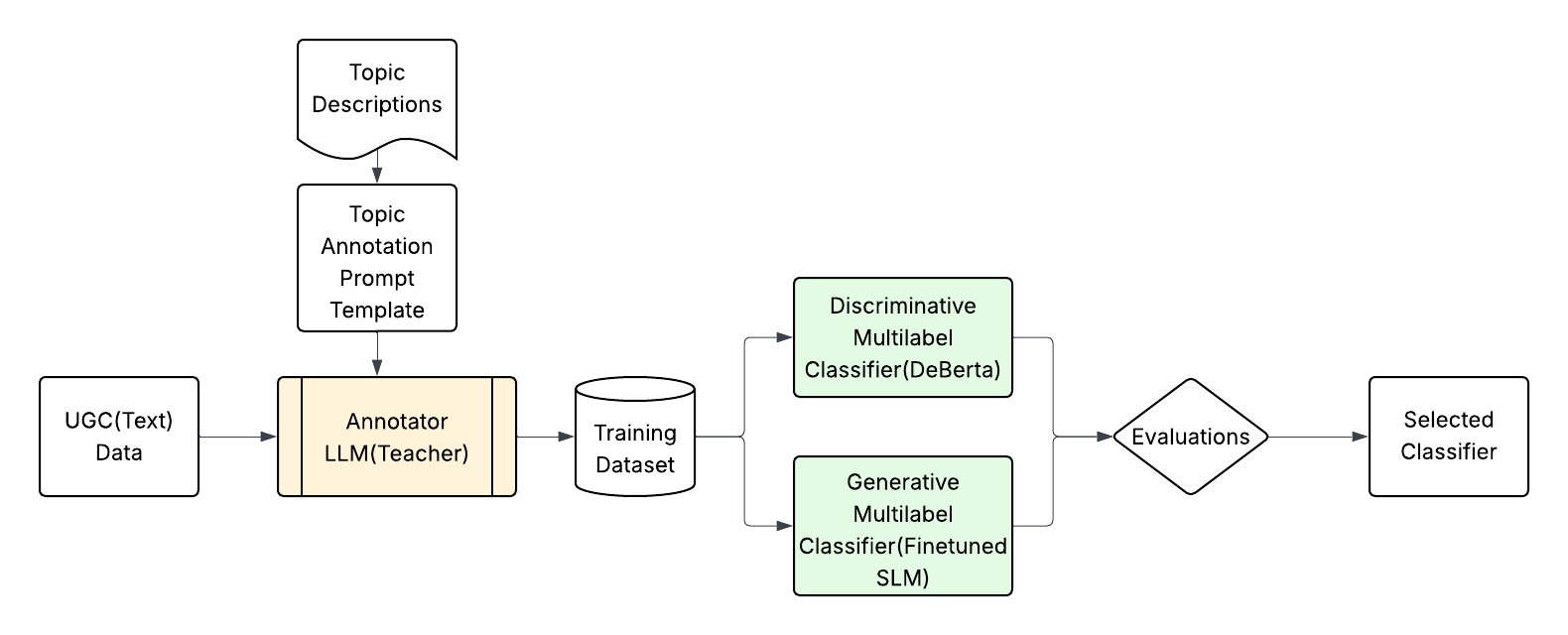}
\caption{Overview of the Scalable Multi-Label Classification Methodology.}
\label{fig:methodology_overview}
\end{figure}

The task of topic assignment within the marketplace ecosystem is modeled as a large-scale multi-label classification problem. Our proposed methodology for constructing these models is illustrated in Figure~\ref{fig:methodology_overview}.

To overcome the significant bottleneck of manual data labeling, we employ a knowledge distillation framework using a Teacher-Student architecture. In this setup, a high-capacity Large Language Model (LLM) acts as a labeling oracle, generating ground truth annotations for a corpus of unlabeled user-generated content (UGC). The oracle is conditioned on a structured taxonomy consisting of class definitions and disambiguation criteria. By leveraging LLMs to synthesize ground truth, we overcome the labeling bottlenecks typical of niche e-commerce categories---a strategy that has shown significant promise in scaling extreme multi-label classification tasks \citep{generativeai_multilabel2024}. We subsequently train smaller, latency-optimized ``student'' models on this distilled ground truth to satisfy the requirements of high-traffic production infrastructure.

\subsection{Student Model Architectures}

To evaluate the most effective architecture for production topic assignment, we compare two distinct modeling paradigms:

\subsubsection{Discriminative Architecture (DeBERTa-V3)}

Traditionally, classification pipelines for user-generated text have relied on discriminative architectures. In this approach, we treat multi-label classification as a series of $N$ independent binary tasks. The model processes input text and passes the final hidden state through a linear layer with a Sigmoid activation function for each potential topic. We formulate this mathematically by optimizing the Binary Cross-Entropy (BCE) loss:
\begin{equation}
\begin{aligned}
\label{eq:bce}
 \mathcal{L}_{BCE} = &\sum_{i=1}^{N} \Big[ y_i \log(\sigma(W_i^T h)) \\
  & + (1-y_i) \log(1 - \sigma(W_i^T h)) \Big]
 \end{aligned}
\end{equation}
where $h$ is the final hidden state representation, $W_i$ represents the topic-specific weights, and $y_i \in \{0,1\}$ is the ground truth label. Systems like Text2Topic have demonstrated the efficacy of such discriminative approaches in transforming millions of unstructured reviews into structured business insights \citep{text2topic2023}. However, by treating each label as an isolated boundary, this architecture often fails to capture the semantic interplay between co-occurring topics.

\subsubsection{Generative Architecture (Fine-tuned SLMs)}

In contrast, our generative approach leverages the autoregressive nature of Small Language Models (SLMs). Instead of predicting a fixed set of logits, the model is fine-tuned to sequentially generate tokens representing the relevant topics. The generative approach models topic assignment as a sequence-to-sequence autoregressive decoding task, conditioned on the input text $X$ and Low-Rank Adaptation (LoRA) weights~\citep{hu2021lora} $\theta_{LoRA}$:
\begin{equation}
    P(Y|X) = \prod_{t=1}^{T} P(y_t | y_{<t}, X, \theta_{LoRA})
\end{equation}
where $Y$ is the generated sequence of topic tokens. While traditional classifiers treat labels as independent targets, generative models can better reason through the label space by capturing semantic relationships between topics as they are decoded \citep{llm_multilabel2025}. Furthermore, the generative architecture is independent of the total number of labels; this flexibility allows the system to adapt to evolving taxonomies without requiring a full retraining of the output head every time a new class is introduced.

\textbf{Architectural Trade-offs:} In summary, discriminative baselines offer lower inference latency and are highly optimized for fixed $N$-way binary classification. However, their reliance on independent sigmoid heads prevents them from capturing semantic co-occurrence and makes taxonomy expansion rigid, as new classes require retraining the output head. Conversely, while generative SLMs incur higher inference latency, their sequence-to-sequence autoregressive decoding natively captures label dependencies and provides high taxonomy flexibility without architectural modifications.

\subsection{LLM-Based Ground Truth Generation}

In a large-scale marketplace, user-generated content often results in vast repositories of unlabeled historical data. To address the prohibitive cost of manual annotation, we utilize an external LLM (GPT-4o) as a labeling oracle to generate high-fidelity synthetic ground truth.

\subsubsection{Generalized Prompt Template}

To ensure consistency across domains, we developed a generalized template instantiated with domain-specific taxonomies and exemplars. The core prompt structure is outlined below:

\begin{quote}
\small
\ttfamily
SYSTEM ROLE:\\
You are an AI assistant specialized in analyzing [DOMAIN] and identifying relevant topics from a predefined taxonomy. Your task is to classify the given text and return all applicable topic labels.\\

INSTRUCTIONS:\\
1. Analyze the input text carefully\\
2. Identify ALL relevant topics from the taxonomy below\\
3. A single input can have multiple topics\\
4. Only select topics that clearly apply\\
5. Return your response in valid JSON format\\

TOPIC TAXONOMY:\\
- Topic-1: [Clear definition of what this topic represents, including criteria for assignment, distinguishing features and edge conditions.]\\
- Topic-N: [Complete taxonomy of all possible labels]\\

OUTPUT FORMAT:\\
Return a JSON object with this exact structure:\\
\{\\
\indent "reasoning": "Brief explanation of why these topics apply",\\
\indent "labels": ["Topic-1", "Topic-2"]\\
\}\\

FEW-SHOT EXAMPLES:\\
Input: [Example text 1]\\
Output: \{"reasoning": "Explanation", "labels": ["Topic-1"]\}\\

INPUT TEXT:\\
\{input\_text\}
\end{quote}

This prompt requires a structured taxonomy...

\subsubsection{Prompt Tuning for Quality Assurance}

A critical limitation of this methodology is that the accuracy of the fine-tuned student model is constrained by the quality of the ground truth provided by the oracle. Consequently, generating high-fidelity labels is an iterative process requiring multiple rounds of prompt tuning and validation exercises to minimize hallucinations and ensure taxonomic alignment. Each cycle involved manual audits by domain experts to verify that the distilled signal was robust across both head and long-tail categories, ensuring the student model inherited a high-quality training signal suitable for production.

\section{Experimental Setup and Use Cases}
\label{sec:experimental-setup}

We validate our proposed methodology across two distinct use cases representing different facets of textual user-generated content at a global marketplace. All experiments were conducted using an internal data orchestration and AutoML pipeline, which provides an integrated platform for data fetching, LLM-based data annotation, and model training.

\subsection{Use Case 1: Buyer-Seller Conversational Topic Extraction}
\label{sec:usecase1}

Buyer-seller conversations contain critical e-commerce signals related to price negotiations, shipping inquiries, item questions, and returns. The objective of this use case is to automatically identify these signals in multi-turn dialogues to power downstream automation systems.  

To capture the vast semantic range of the end-to-end buyer-seller journey, we define a comprehensive taxonomy of 112 distinct conversational topics. Instead of relying on broad categorizations, this schema enables fine-grained, multi-label understanding by spanning top-level domains such as Pricing \& Offers, Shipping \& Delivery, Returns \& Refunds, Payments, and Item Issues.  

The initial dataset comprises raw, individual chat logs. To reconstruct the multi-turn conversational flow, these isolated messages were chronologically aggregated using unique thread identifiers. Furthermore, because the raw logs do not explicitly label participant roles, we deterministically resolved buyer and seller identities by explicitly adding them in the conversation. To capture the evolving nature of conversations, we employed a progressive context strategy: for a conversation with $N$ messages, we generate $N$ training examples where each $i$-th turn is classified conditioned on the preceding $i-1$ turns as conversational context.

\begin{quote}
\small
\textbf{Example: Progressive Context Generation} \\
\textbf{Turn 1 Input:} [Buyer] ``Can I pick it up locally?'' \\
\textbf{Turn 1 Label:} \textit{Pickup-Availability} \\[4pt]
\textbf{Turn 2 Input:} [Buyer] ``Can I pick it up locally?'' $\rightarrow$ [Seller] ``Yes, pickup is available! Where are you located?'' \\
\textbf{Turn 2 Label:} \textit{Pickup-Availability-Confirmation, Pickup-Location}
\end{quote}

Because e-commerce intents naturally exhibit extreme class imbalance, many specialized topics appear infrequently. To address this long-tail data scarcity, we leveraged embedding-based similarity—using dense retrieval against a small set of example queries—to source an additional 10,000 conversations from historical logs specifically rich in rare topics. The final distilled dataset comprises of about 25,000 training conversations (~135K messages) and a held-out test set of ~7,600 conversations (~40K messages)

\subsection{Use Case 2: Product Review Topic Classification}
\label{sec:prodreviewusecase}

Product reviews are a critical trust signal for buyers, but generic reviews often fail to address category-specific concerns, limiting their utility for complex purchasing decisions. To improve review comprehensiveness, we developed a topic identification system that provides real-time, category-specific cues as users compose a feedback. For this study, we focus on a subset of five distinct product categories: Tires, Parts \& Accessories, Lipstick, Vacuum Cleaners, and Juicers. Rather than capturing generic sentiment, the taxonomy isolates category-specific evaluation dimensions. For example, for Tires category the topics consists of Tread Wear, Traction; Yield and Cleanup for Juicers; and Pigment and Transfer for Lipstick. We have about 4-5 topics for a category(total 21 categories).

A core engineering challenge here is scalability: building a single model that generalizes across thousands of product categories—without requiring isolated per-category classification heads—while maintaining strict real-time inference latency. We model this as an extreme multi-label classification task over a large taxonomy. The dataset consists of historical user-generated product reviews collected from listings across these categories. Reviews were strictly filtered to retain only those with substantial textual content (a minimum of 100 characters). Following LLM-based ground-truth distillation, the final review dataset consists of approximately 39,000 training examples and 6,800 test examples aggregated across the target categories. For ground truth, we instantiated our generalized prompt template with taxonomy specific to a category, to ensure the student models inherited a high-fidelity training signal.

\begin{table*}[t]
\centering
\begin{tabular}{lcccc}
\hline
\textbf{Model Architecture} & \textbf{Parameters} & \textbf{Reviews (Macro-F1)} & \textbf{Conversations (Macro-F1)} & \textbf{P95 Latency (ms)} \\
\hline
Discriminative (DeBERTa-V3) & 186M & 0.7047 & 0.4062 & \textbf{< 100} \\
Generative (SLM-1B) & 1B & 0.6502 & 0.4501 & 163 \\
Generative (SLM-4B) & 4B & 0.7405 & 0.4908 & 208 \\
Generative (SLM-8B) & 8B & \textbf{0.8031} & \textbf{0.5404} & 241 \\
\hline
\end{tabular}
\caption{Master Performance and Latency Comparison Across Architectures}
\label{tab:master_results}
\end{table*}

\begin{table*}[t]
\centering
\begin{tabular}{lccccc}
\toprule
\textbf{Training Support} & \textbf{DeBERTa-V3} & \textbf{ModernBERT} & \textbf{SLM-8B} & \textbf{SLM-4B} & \textbf{SLM-1B} \\
 & \textbf{(Macro)} & \textbf{(Macro)} & \textbf{(Macro)} & \textbf{(Macro)} & \textbf{(Macro)} \\
\midrule
$\ge 1$ & 0.350 & 0.304 & 0.473 & 0.457 & 0.436 \\
$\ge 10$ & 0.350 & 0.304 & 0.475 & 0.461 & 0.440 \\
$\ge 50$ & 0.357 & 0.309 & 0.500 & 0.482 & 0.461 \\
$\ge 100$ & 0.376 & 0.330 & 0.527 & 0.513 & 0.489 \\
$\ge 200$ & 0.407 & 0.354 & 0.572 & 0.556 & 0.528 \\
$\ge 250$ & 0.429 & 0.373 & 0.578 & 0.559 & 0.536 \\
$\ge 500$ & 0.474 & 0.420 & 0.596 & 0.579 & 0.557 \\
$\ge 1000$ & 0.506 & 0.458 & 0.602 & 0.585 & 0.565 \\
\bottomrule
\end{tabular}
\caption{Macro-F1 performance metrics on Conversational dataset across different training support levels.}
\label{tab:m2m_support}
\end{table*}

\begin{table*}[t]
\centering
\begin{tabular}{lccccc}
\toprule
\textbf{Training Support} & \textbf{DeBERTa-V3} & \textbf{ModernBERT} & \textbf{SLM-8B} & \textbf{SLM-4B} & \textbf{SLM-1B} \\
 & \textbf{(Macro)} & \textbf{(Macro)} & \textbf{(Macro)} & \textbf{(Macro)} & \textbf{(Macro)} \\
\midrule
$\ge 1$ & 0.670 & 0.687 & 0.796 & 0.653 & 0.646 \\
$\ge 200$ & 0.670 & 0.687 & 0.796 & 0.653 & 0.646 \\
$\ge 250$ & 0.703 & 0.690 & 0.798 & 0.652 & 0.646 \\
$\ge 500$ & 0.742 & 0.718 & 0.811 & 0.642 & 0.643 \\
$\ge 1000$ & 0.749 & 0.731 & 0.814 & 0.633 & 0.635 \\
\bottomrule
\end{tabular}
\caption{Macro-F1 performance metrics on Product Review dataset across different training support levels.}
\label{tab:product_review_support}
\end{table*}

\subsection{Model Configurations}
We evaluated two distinct architectural paradigms on the distilled datasets:

\paragraph{Discriminative Models (DeBERTa-V3 and ModernBERT)}
We fine-tuned DeBERTa-V3-base\citep{DeBERTaV3} and ModernBERT\citep{warner2024smarterbetterfasterlonger} using a multi-label classification head with independent sigmoid activations for each topic. Models were trained for 25 epochs with a batch size of 16, using the AdamW optimizer with a learning rate of 5e-5, 1,000 warmup steps, and a weight decay of 0.01. We used the default values for all other hyperparameters.

\paragraph{Generative Models (SLM-1B, SLM-4B, SLM-8B)}
The generative models consists of a family of decoder-only Small Language Models(SLMs) with 1 billion, 4 billion, and 8 billion parameters. All three models share a common transformer architecture utilizing grouped-query attention (GQA), RoPE positional embeddings, SwiGLU activations, untied input/output embeddings, and a shared 130k-token vocabulary. The model scales are configured as follows:
\begin{itemize}
    \item \textbf{SLM-1B:} 28 layers, hidden size 1,536, 12 attention heads (2 GQA groups), and a 32k-token context window.
    \item \textbf{SLM-4B:} 36 layers, hidden size 3,072, 24 attention heads (8 GQA groups), and an 8k-token context window.
    \item \textbf{SLM-8B:} 32 layers, hidden size 4,096, 32 attention heads (8 GQA groups), and a 32k-token context window.
\end{itemize}

These models were parameter-efficiently fine-tuned for sequence-to-sequence topic generation using a distributed multi-LoRA (Low-Rank Adaptation) infrastructure. LoRA was applied to all projection layers with a rank of 64 and an alpha of 128. The SLMs were trained for 3 epochs with a learning rate of 1e-4 using a linear schedule, a dropout of 0.05, and a maximum token length of 4,096.

Both model families were subsequently evaluated on held-out test sets to measure Macro-F1 fidelity, inference latency, and scalability characteristics.

\section{Results and Discussion}
\label{sec:results}

\begin{table}[t]
\centering
\resizebox{\columnwidth}{!}{%
\begin{tabular}{lccc}
\toprule
\textbf{Dataset} & \textbf{Expansion} & \textbf{DeBERTa-V3} & \textbf{SLM-8B} \\
\midrule
Conversations & 10 $\to$ 112 Labels & -35.53\% & \textbf{-0.23\%} \\
Product Reviews & 4 $\to$ 21 Labels & -15.01\% & \textbf{-4.46\%} \\
\bottomrule
\end{tabular}%
}
\caption{F1 Macro performance change with label expansion.}
\label{tab:label_scale}
\end{table}

We evaluate the performance of our generative multi-label models (SLM-1B, SLM-4B, SLM-8B) against two discriminative baselines (DeBERTa-V3 and ModernBERT) across four critical dimensions: dataset complexity, training support, label scale, and inference latency. Following established best practices for skewed multi-label distributions, we prioritize the Macro-F1 score to ensure the model's ability to handle long-tail, infrequent labels is accurately captured.

At peak parameter scale (SLM-8B), the generative architecture outperformed all discriminative baselines across both datasets. Between the two discriminative candidates, DeBERTa-V3 consistently demonstrated superior performance compared to ModernBERT. To maintain clarity, we subsequently refer to DeBERTa-V3 as the primary baseline. Similarly, to streamline our taxonomy expansion and scalability analyses, we benchmark this primary baseline specifically against our highest-performing generative model (SLM-8B) in those subsequent sections.

\subsection{Dataset Complexity (The Data-Dependent Trade-off)}
The first dimension evaluates how model architectures handle different linguistic structures. As detailed in Table~\ref{tab:master_results}, we observed a critical performance divergence based on the target data.

For structured, single-author Product Reviews, the discriminative DeBERTa-V3 model (0.7047) outperforms the ultra-lightweight generative SLM-1B (0.6502). This indicates that when intents are concentrated, traditional classification boundaries are highly efficient. However, scaling the generative model unlocks superior semantic reasoning, allowing SLM-8B to dominate the category (0.8031).

Conversely, in the fluid, informal, and overlapping intents of multi-turn conversational data, the gap is substantially wider. In this noisy environment, even the smallest 1B-parameter generative model (0.4501) surpasses the discriminative baseline (0.4062). The autoregressive nature of SLMs allows them to better navigate the context shifts typical of member dialogues.

\subsection{Training Support (Handling Data Scarcity)}
In e-commerce, rare labels (the ``long tail'') often lack sufficient training data. We tested model performance as the minimum number of training examples per label (Support) varied Table~\ref{tab:m2m_support} and Table~\ref{tab:product_review_support}.

At the lowest data density (Support $\ge$1), the generative SLM-8B model outperforms the best discriminative baseline by over 12.00 percentage across both datasets. The generative model leverages its vast pre-training to handle rare labels where task-specific data is missing. It understands the semantic meaning of a topic rather than just seeing it as a statistical class index, resulting in a significantly higher baseline performance and data efficiency. In contrast, discriminative models show a much steeper climb, requiring a massive increase in label volume to approach the performance of the generative models.

\subsection{Label Scale (Robustness to Taxonomy Growth)}
This dimension measures how performance degrades as we increase the number of labels the model must identify simultaneously, simulating the real-world scenario of expanding category coverage.

As shown in Table~\ref{tab:label_scale}, the discriminative model suffered a catastrophic collapse (-35.53\% Macro-F1) in the conversational dataset as the label space expanded. Because discriminative models treat multi-labeling as isolated binary tasks, the model struggles to define distinct boundaries as the label space crowds. Conversely, the generative model demonstrated extreme robustness (-0.23\% drop), maintaining its performance even as the target taxonomy scaled to 112 concurrent topics.
\begin{table}[htbp]
\centering
\begin{tabular}{lcc}
\toprule
\textbf{Label Configuration} & \textbf{DeBERTa-V3} & \textbf{SLM-8B} \\
 & \textbf{Macro-F1} & \textbf{Macro-F1} \\
\midrule
10 Random Labels & 0.5435 & 0.4739 \\
10+15 Labels & 0.5132 & 0.5553 \\
25+25 Random Labels & 0.4890 & 0.5824 \\
All Random Labels (112) & 0.3504 & 0.4728 \\
\bottomrule
\end{tabular}
\caption{Macro-F1 performance with varying label configurations on Conversational dataset.}
\label{tab:scalability_m2m}
\end{table}

\begin{table}[htbp]
\centering
\begin{tabular}{lcc}
\toprule
\textbf{Category Configuration} & \textbf{DeBERTa-V3} & \textbf{SLM-8B} \\
 & \textbf{Macro-F1} & \textbf{Macro-F1} \\
\midrule
Tires Only & 0.7879 & 0.8334 \\
Tires and P\&A & 0.6987 & 0.7924 \\
All 5 Categories & 0.6696 & 0.7962 \\
\bottomrule
\end{tabular}
\caption{Macro-F1 performance with varying category combinations on Product Review dataset.}
\label{tab:scalability_product_review}
\end{table}

\subsection{Inference Latency and Production Feasibility}
While SLM-8B achieves the highest theoretical accuracy across all dimensions, its 241ms P95 latency is significantly heavier than traditional encoder-only discriminative models. However, our scaling analysis proves that practitioners facing strict hardware constraints do not have to abandon generative reasoning. By deploying SLM-1B, engineering teams can achieve a 32\% reduction in latency (163ms) compared to the 8B model while still maintaining superior classification fidelity over discriminative baselines on complex, unstructured conversational data.

\begin{table*}[t]
\centering
\begin{tabular}{llcc}
\toprule
\textbf{Architecture} & \textbf{Model} & \textbf{Latency (ms)} & \textbf{Relative to DeBERTa-V3} \\
\midrule
\multirow{2}{*}{Discriminative} & DeBERTa-V3 & \textbf{58}  & 1.0$\times$ \\
                                & ModernBERT  & 79           & 1.4$\times$ \\
\midrule
\multirow{3}{*}{Generative}     & SLM-1B      & 163          & 2.8$\times$ \\
                                & SLM-4B      & 208          & 3.6$\times$ \\
                                & SLM-8B      & 241          & 4.2$\times$ \\
\bottomrule
\end{tabular}
\caption{Inference latency comparison between discriminative and generative model families}
\label{tab:model-latency}
\end{table*}

\subsection{Out-of-Vocabulary Labels}
Table \ref{tab:oov_labels} summarizes the tendency of generative models to generate labels outside the fixed training label set. Overall, OOV generation is rare at the sample level: even the largest 8B model produces OOV labels for only 100 out of 39,172 evaluation examples, corresponding to just 0.26\% of the test set.

However, the diversity of these OOV labels is non-negligible. The 1B, 4B, and 8B models generate 22, 17, and 29 unique OOV labels, respectively. This suggests that larger generative models do not necessarily become more constrained to the observed label space; instead, the 8B model produces the largest number of distinct OOV labels.

A qualitative inspection reveals that most OOV labels are semantically plausible rather than arbitrary strings. For example, frequent OOV predictions such as \textit{shipping-issue}, \textit{shipping-label}, and \textit{return-issue} resemble valid intent labels, combining known domain concepts with reasonable modifiers. This behavior indicates that the models are leveraging their generative capacity to synthesize label-like outputs, especially in high-frequency domains like shipping and returns.

While this semantic flexibility is impressive, it is undesirable for closed-set classification because such outputs cannot be directly mapped to the predefined taxonomy. To mitigate this in our production deployment, we apply a lightweight post-processing normalization step that either maps these high-frequency OOV variants to their nearest taxonomic equivalent via semantic similarity, or safely discards them to strictly enforce the schema constraint.

\begin{table*}[t]
\centering
\small
\begin{tabular}{lrrrrl}
\hline
\textbf{Model} &
\textbf{\# OOV} &
\textbf{OOV label rate} &
\textbf{OOV samples} &
\textbf{Sample rate} &
\textbf{Most frequent OOV labels} \\
\hline
SLM-1B & 22 & 19.6\% & 42 / 39,172 & 0.11\% &
\texttt{shipping-issue}, \texttt{shipping-insurance} \\
SLM-4B & 17 & 15.2\% & 54 / 39,172 & 0.14\% &
\texttt{shipping-complaint}, \texttt{shipping-issue} \\
SLM-8B & 29 & 25.9\% & 100 / 39,172 & 0.26\% &
\texttt{shipping-issue}, \texttt{shipping-label} \\
\hline
\end{tabular}
\caption{Out-of-vocabulary (OOV) labels produced by generative models.
The OOV label rate is computed against the 112 labels observed during training, while the sample rate measures the fraction of evaluation samples for which the model generated an OOV label.}
\label{tab:oov_labels}
\end{table*}

\section{Impact and Production Deployment}
\label{sec:impact}
While our offline evaluation demonstrated the architectural advantages of generative models across both structured and unstructured domains, deploying LLMs to live marketplace traffic introduces significant engineering constraints. We successfully deployed the optimized SLM-8B model for both primary use cases to production.
For the conversational topic classification system (Section 3.1), processing a throughput of approximately 250 requests per second required a cluster of 9 H100 GPUs to maintain a P99 latency of under 500 milliseconds.
For the Product Review domain (Section 3.2), the model was launched for the Tires and Automotive Parts \& Accessories categories, also utilizing H100 GPU infrastructure. Across both applications, the generative deployments successfully achieved a P95 latency of less than 400 milliseconds. Crucially, for the product review deployment, integrating this model into the user-facing "Suggested Topic" interface allowed us to provide real-time, category-specific writing cues to buyers. An online A/B test conducted during a 20\% traffic ramp-up demonstrated that this offline generative reasoning directly translates to platform health, yielding a 60\% increase in the volume of product reviews meeting the platform's strict quality thresholds for approval.

\section{Conclusion and Future Directions}
\label{sec:conclusion}

This paper presented a comprehensive architectural comparison for large-scale multi-label topic assignment, evaluating both discriminative baselines and generative Small Language Models (SLMs) within a production Teacher-Student distillation framework. To overcome the manual annotation bottleneck, we leveraged high-capacity LLMs as labeling oracles to synthesize high-fidelity training data across multiple e-commerce domains.
Our empirical analysis revealed a critical, data-dependent architectural trade-off. While optimized discriminative models (DeBERTa-V3) maintain an efficiency edge for structured, single-author text like product reviews, generative sequence-to-sequence classification offers vastly superior reasoning for complex, unstructured data. In noisy, multi-turn conversational environments, generative SLMs surpassed discriminative baselines by a massive relative margin. Furthermore, generative models demonstrated near-total immunity to taxonomy expansion, maintaining robust F1 scores where traditional classification boundaries suffered a severe 35.53\% collapse. By successfully deploying these optimized models on GPU infrastructure and achieving strict sub-400ms P95 latencies, we demonstrated that offline generative reasoning translates directly to platform health, yielding a 60\% increase in approved review volume during live A/B testing.
Moving forward, our research will focus on dynamic taxonomy expansion. We aim to explore whether fine-tuned generative SLMs can consistently identify new, unseen topics via semantic prompt descriptions in a completely ``zero-shot'' manner, thereby bypassing the need for task-specific retraining as marketplace categories continuously evolve.
\bibliography{references}

\end{document}